\documentclass{article} 
\usepackage{iclr2022_conference,times}

\usepackage{amsmath,amsfonts,bm}

\def\eqref#1{equation~\ref{#1}}

\def\1{\bm{1}}

\DeclareMathAlphabet{\mathsfit}{\encodingdefault}{\sfdefault}{m}{sl}
\SetMathAlphabet{\mathsfit}{bold}{\encodingdefault}{\sfdefault}{bx}{n}

\newcommand{\Ni}{({\em i})~}
\newcommand{\Nii}{({\em ii})~}
\newcommand{\Niii}{({\em iii})~}

\usepackage{hyperref}
\usepackage{url}
\usepackage[utf8]{inputenc} 
\usepackage[T1]{fontenc}    
\usepackage{booktabs}       
\usepackage{amsfonts}       
\usepackage{nicefrac}       
\usepackage{microtype}      
\usepackage{xcolor}         
\usepackage{amsmath}
\usepackage{graphicx}
\usepackage{subfigure}
\usepackage{multirow}
\usepackage{amsthm}
\usepackage{wrapfig}
\usepackage{color}

\def\ff{\text{ff}}
\newtheorem{definition}{Definition}
\newtheorem{theorem}{Theorem}
\newtheorem{lemma}[theorem]{Lemma}

\newtheorem{innercustomgeneric}{\customgenericname}
\providecommand{\customgenericname}{}
\newcommand{\newcustomtheorem}[2]{%
  \newenvironment{#1}[1]
  {%
   \renewcommand\customgenericname{#2}%
   \renewcommand\theinnercustomgeneric{##1}%
   \innercustomgeneric
  }
  {\endinnercustomgeneric}
}

\newcustomtheorem{customthm}{Theorem}
\newcustomtheorem{customlemma}{Lemma}

\title{Revisiting Over-smoothing in BERT from the Perspective of Graph}

\author{Han Shi\textsuperscript{\rm 1}\thanks{Equal contribution.}, Jiahui Gao\textsuperscript{\rm 2}\footnotemark[1], Hang Xu\textsuperscript{\rm 3}, Xiaodan Liang\textsuperscript{\rm 4}, \\ \textbf{Zhenguo Li\textsuperscript{\rm 3}, Lingpeng Kong\textsuperscript{\rm 2}, Stephen M.S. Lee\textsuperscript{\rm 2}, James T. Kwok\textsuperscript{\rm 1}}\\
  \textsuperscript{\rm 1}Hong Kong University of Science and Technology,
  \textsuperscript{\rm 2}The University of Hong Kong, \\
  \textsuperscript{\rm 3}Huawei Noah’s Ark Lab,
  \textsuperscript{\rm 4}Sun Yat-sen University \\
  \texttt{\{hshiac,jamesk\}@cse.ust.hk,\{sumiler,smslee\}@hku.hk,lpk@cs.hku.hk}, \\
  \texttt{\{xu.hang,li.zhenguo\}@huawei.com,xdliang328@gmail.com} \\
}

\iclrfinalcopy 
\begin{document}

\maketitle
\begin{abstract}
Recently over-smoothing phenomenon of Transformer-based models is observed in both vision and language fields.
However, no existing work has delved deeper to further investigate the main cause
of this phenomenon.
In this work, we make the attempt to analyze the over-smoothing problem from the perspective of graph, where such problem was first discovered and explored. Intuitively,
the self-attention matrix can be seen as a normalized adjacent matrix of a corresponding graph. Based on the above connection, we provide some theoretical analysis and find that layer normalization plays a key role in the over-smoothing issue of Transformer-based models. Specifically, if the standard deviation of layer normalization is sufficiently large, the output of Transformer stacks will converge to a specific low-rank subspace and result in over-smoothing.
To alleviate the over-smoothing problem, we consider hierarchical fusion strategies, which combine the representations from different layers adaptively to make the output more diverse. Extensive experiment results on various data sets illustrate the effect of our fusion method.
\end{abstract}

\section{Introduction}
Over the past few years, Transformer \citep{vaswani2017attention} has been widely used
in various natural language processing (NLP) tasks,
including
text classification \citep{wang2018glue}, text translation \citep{ott2018scaling},
question answering \citep{rajpurkar2016squad,rajpurkar2018know} and text generation \citep{brown2020language}. The recent application of Transformer in computer vision (CV) field also demonstrate the potential capacity of Transformer architecture. For instance, Transformer variants have been successfully used for image classification \citep{dosovitskiy2020image}, object detection \citep{carion2020end} and semantic segmentation \citep{strudel2021segmenter}.
Three fundamental descendants from Transformer include
BERT \citep{devlin2019bert}, RoBERTa \citep{liu2019roberta} and ALBERT \citep{lan2019albert},
which
achieve state-of-the-art performance
on a wide range of NLP tasks.


Recently, \cite{dong2021attention} observes the ``token uniformity'' problem,
which reduces the capacity of
Transformer-based architectures
by making all token representations identical.
They claim that pure self-attention (SAN) modules cause token uniformity,
but they do not discuss whether the token uniformity problem still exists in Transformer blocks.
On the other hand,
\cite{gong2021improve} observe
the ``over-smoothing'' problem
for ViT \citep{dosovitskiy2020image},
in
that different input patches are mapped to a similar latent representation.
To prevent loss of information, they introduce additional loss functions to encourage diversity and successfully improve model performance by suppressing over-smoothing.
Moreover, ``overthinking'' phenomenon,  indicating that shallow representations are better than deep representations, also be observed in  \citep{zhou2020bert,kaya2019shallow}. As discussed in Section~\ref{sec:exist}, this phenomenon has some inherent connection with over-smoothing.
In this paper, we use ``over-smoothing'' to unify the above issues, and refer
this as the phenomenon that the model performance is deteriorated because different inputs are mapped to a similar representation.

As the over-smoothing problem is first studied in the graph neural network (GNN) literature
\citep{li2018deeper,xu2018representation,zhao2019pairnorm}, in this paper, we attempt to explore the cause of such problem by building
a relationship between Transformer blocks and graphs. Specifically, we consider
the self-attention matrix as the normalized adjacency matrix of a weighted graph, whose nodes are the tokens in a sentence.
Furthermore, we consider the inherent connection between BERT and graph convolutional networks \citep{kipf2017semi}. Inspired by the over-smoothing problem in GNN,
we study over-smoothing in BERT from a theoretical view via matrix projection.
As opposed to
\cite{dong2021attention}, where the authors claim that layer normalization is irrelevant to over-smoothing, we
find that layer normalization \citep{ba2016layer} plays an important role in
over-smoothing. Specifically, we theoretically prove that, if the standard deviation in layer normalization is sufficiently
large, the outputs of the Transformer stacks will converge to a low-rank subspace,
resulting in over-smoothing. Empirically, we verify that the conditions hold for a certain number of samples for a pre-trained and fine-tuned BERT model \citep{devlin2019bert}, which is consistent with our above observations.

To alleviate the over-smoothing problem, we propose a hierarchical fusion strategy
that adaptively fuses representations from different layers.
Three fusion approaches are used: \Ni Concat Fusion, \Nii Max Fusion, and \Niii Gate Fusion.
The proposed method reduces the similarity between tokens and outperforms BERT baseline on the GLUE \citep{wang2018glue}, SWAG \citep{zellers2018swag} and SQuAD \citep{rajpurkar2016squad,rajpurkar2018know} data sets.

In summary, the contributions of this paper are as follows: \Ni We develop the
relationship between self-attention and graph for a better understanding of
over-smoothing in BERT. \Nii We provide theoretical analysis on over-smoothing in
the BERT model, and empirically verify the theoretical results. \Niii We propose
hierarchical fusion strategies that adaptively combine different layers to alleviate
over-smoothing. Extensive experimental results verify our methods' effectiveness.

\section{Related Work}
\subsection{Transformer Block and Self-Attention \label{sec:trans_block}}
Transformer block is a basic component in Transformer model \citep{vaswani2017attention}.
Each Transformer block consists of a self-attention layer and a feed-forward layer.
Let $\boldsymbol{X}\in\mathbb{R}^{n\times d}$ be the input to a Transformer block, where $n$ is the number of input tokens and $d$ is the embedding size.
The self-attention
layer output
can be written
as:
\begin{align}
Attn(\boldsymbol{X}) &=  \boldsymbol{X}+\sum_{k=1}^{h}\sigma(\boldsymbol{X}\boldsymbol{W}_{k}^{Q}(\boldsymbol{X}\boldsymbol{W}_{k}^{K})^{\top} )\boldsymbol{XW}_{k}^{V}\boldsymbol{W}_{k}^{O\top}
=\boldsymbol{X}+\sum_{k=1}^{h}\hat{\boldsymbol{A}}_{k}\boldsymbol{X}\boldsymbol{W}_{k}^{VO},  \label{eq:attn}
\end{align}
where $h$ is the number of heads,
$\sigma$ is
the softmax function, and
$\boldsymbol{W}_{k}^{Q},\boldsymbol{W}_{k}^{K},\boldsymbol{W}_{k}^{V},\boldsymbol{W}_{k}^{O}\in\mathbb{R}^{d\times
d_{h}}$ (where $d_{h}=d/h$ is the dimension of a single-head output) are weight matrices for the query, key, value, and output,
respectively of the $k$th head. In particular,
the self-attention matrix
\begin{equation} \label{eq:matrix}
\hat{\boldsymbol{A}}=\sigma(\boldsymbol{XW}^{Q}(\boldsymbol{XW}^{K})^{\top})=\sigma(\boldsymbol{QK}^{\top})
\end{equation}
    in (\ref{eq:attn})
plays a key role in the self-attention layer
\citep{park2019sanvis,gong2019efficient,kovaleva2019revealing}.
As in \citep{yun2020n,shi2021sparsebert,dong2021attention}, we drop the scale
product $1/\sqrt{d_h}$ to simplify analysis.

The feed-forward layer usually has
two fully-connected (FC) layers
with residual connection:
\[
FF(\boldsymbol{X})=Attn(\boldsymbol{X})+ReLU(Attn(\boldsymbol{X})\boldsymbol{W}_{1}+\boldsymbol{b}_{1})\boldsymbol{W}_{2}+\boldsymbol{b}_{2},
\]
where
$\boldsymbol{W}_{1}\in\mathbb{R}^{d\times
d_{\ff}},\boldsymbol{W}_{2}\in\mathbb{R}^{d_{\ff}\times d}$ ($d_{\ff}$ is the size of the intermediate layer)
are the weight matrices, and
$\boldsymbol{b}_{1},\boldsymbol{b}_{2}$
are the biases. Two layer normalization \citep{ba2016layer} operations are
performed after the self-attention layer and fully-connected layer, respectively.

\subsection{Over-smoothing}\label{sec:over-smooth}
In graph neural networks, over-smoothing refers to the problem that the performance deteriorates as representations of all the nodes become similar
\citep{li2018deeper,xu2018representation,huang2020tackling}. Its main cause is
the stacked aggregation layer using the same adjacency matrix. Recently, several approaches have been proposed to alleviate the over-smoothing problem.
\cite{xu2018representation} propose a jumping knowledge network for better structure-aware representation, which flexibly leverages different neighborhood ranges.
ResGCN
\citep{li2019deepgcns}
adapts the residual connection and dilated convolution in
the graph
convolutional network (GCN), and successfully scales the GCN to $56$ layers.
\cite{zhao2019pairnorm}
propose
PairNorm,
a novel normalization layer, that prevents node embeddings from becoming too similar.
DropEdge
\citep{rong2019dropedge,huang2020tackling}
randomly removes edges from the input graph at each training epoch,
and reduces the effect of over-smoothing.

Unlike graph neural networks, over-smoothing in Transformer-based architectures has not been discussed in detail.
\cite{dong2021attention}
introduce
the
``token-uniformity''
problem
for self-attention, and show that skip
connections and multi-layer perceptron can mitigate this problem. However,
\cite{gong2021improve} still observe over-smoothing
on the Vision Transformers \citep{dosovitskiy2020image}.

\section{Does Over-smoothing Exist in BERT?} \label{sec:exist}

\begin{figure}[t]
\centering
\subfigure[Token-wise cosine similarity.\label{subfig:similarity}]{\includegraphics[width=0.33\textwidth]{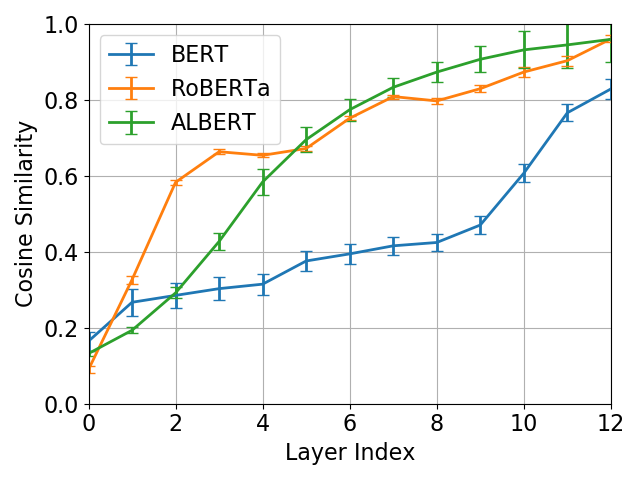}}
\subfigure[Token similarity and error.\label{subfig:overthinking}]{\includegraphics[width=0.33\textwidth]{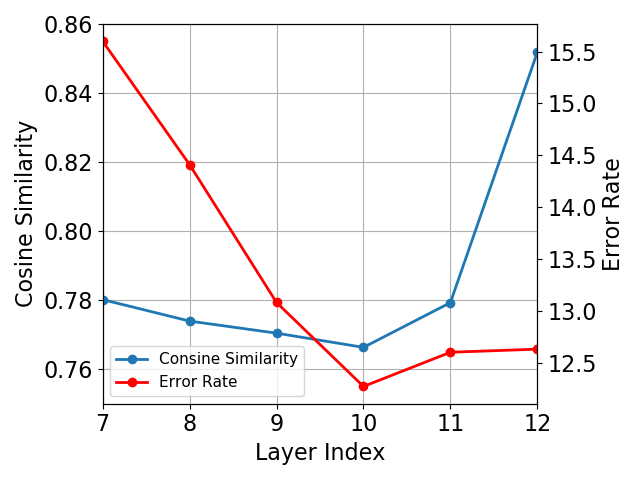}}
\subfigure[Hierarchical fusion.\label{subfig:motivation}]{\includegraphics[width=0.32\textwidth]{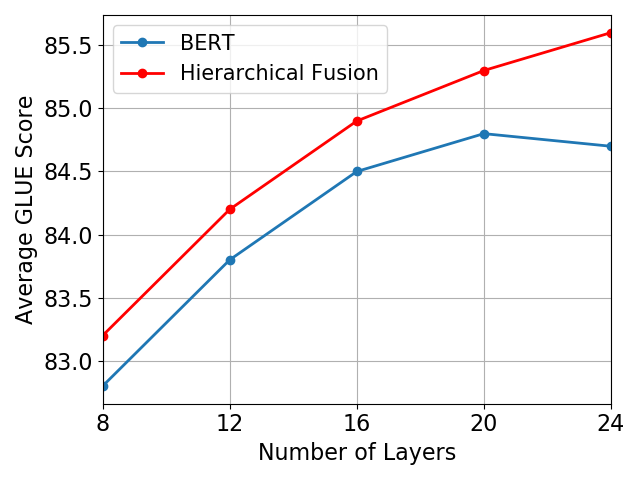}}
\caption{Over-smoothing in BERT models.}
\label{fig:1}
\end{figure}

In this section, we first explore the
existence of
over-smoothing
in BERT, by
measuring the similarity between tokens
in each
Transformer layer.
Specifically, we use the token-wise cosine similarity \citep{gong2021improve} as our similarity measure:
\[ \text{CosSim}=\frac{1}{n(n-1)}\sum_{i\neq j}\frac{\boldsymbol{h}_i^\top
	 \boldsymbol{h}_j}{\Vert \boldsymbol{h}_i\Vert_2\Vert
	 \boldsymbol{h}_j\Vert_2}, \]
where $n$ is the number of tokens, $\boldsymbol{h}_i$ and $\boldsymbol{h}_j$ are
two representations of  different tokens, and $\Vert\cdot\Vert_2$ is the Euclidean norm.
Following \cite{dong2021attention}, we use
WikiBio \citep{lebret2016neural} as input to the following  Transformer-based models fine-tuned on the SQuAD data set \citep{rajpurkar2018know}: \Ni BERT
\citep{devlin2019bert}, \Nii RoBERTa \citep{liu2019roberta} and \Niii ALBERT
\citep{lan2019albert}.\footnote{Our implementation is based on the HuggingFace’s
Transformers library \citep{wolf2020Transformers}.}
For comparison,
all three models are stacked with $12$ blocks.
We calculate each \textit{CosSim} for each data sample and
show
the average and standard derivation of \textit{CosSim} values over all WikiBio data.

In the figures, layer $0$ represents original input token representation, and layer $1$-$12$ represents the corresponding transformer layers.
As shown in Figure~\ref{subfig:similarity},
the original token representations
are different from each
other, while token similarities are high in the last layer. For instance, the average token-wise
cosine similarity of the last layer of ALBERT and RoBERTa are both larger than $90\%$.

To illustrate the relationship between  ``over-thinking''
and ``over-smoothing'',
we compare the token-wise cosine similarity
at each layer
with the corresponding
error rate.
As for the corresponding
error rate of layer $i$, we use the representations from layer $i$ as the final output and fine-tune the classifier.
Following \cite{zhou2020bert}, we experiment with ALBERT \citep{lan2019albert} fine-tuned on the MRPC
data set \citep{dolan2005automatically} and use their error rate results for convenience. As shown in
Figure~\ref{subfig:overthinking},
layer $10$ has the lowest cosine similarity and error rate.  At layers $11$ and
$12$,
the tokens
have larger cosine similarities,
making them harder to distinguish and resulting in the performance drop.
Thus, ``over-thinking'' can be explained by ``over-smoothing''.

A direct consequence of over-smoothing is that the performance cannot be improved
when the model gets deeper,
since the individual tokens are no longer distinguishable.
To illustrate this, we increase the number of layers in BERT  to 24
while keeping the other settings. As shown in
Figure~\ref{subfig:motivation},
the performance of vanilla BERT
cannot improve as
the model gets deeper. In contrast,
the proposed hierarchical fusion
(as will be discussed in Section~\ref{sec:method})
consistently
outperforms
the baseline,
and
has better and better performance
as the
model
gets
deeper.
Based on these observations,
we conclude that the
over-smoothing problem still exists in BERT.

\section{Relationship between Self-Attention and Graph} \label{sec:relationship}

Since over-smoothing is first discussed in the graph neural network
literature \citep{li2018deeper,zhao2019pairnorm},
we attempt to understand its cause from a graph perspective
in this section.

\subsection{Self-Attention vs ResGCN} \label{sec:formulation}

\begin{figure}[t]
\centering
\subfigure[Graph $\mathcal{G}$.\label{subfig:graph}]{\includegraphics[width=0.27\textwidth]{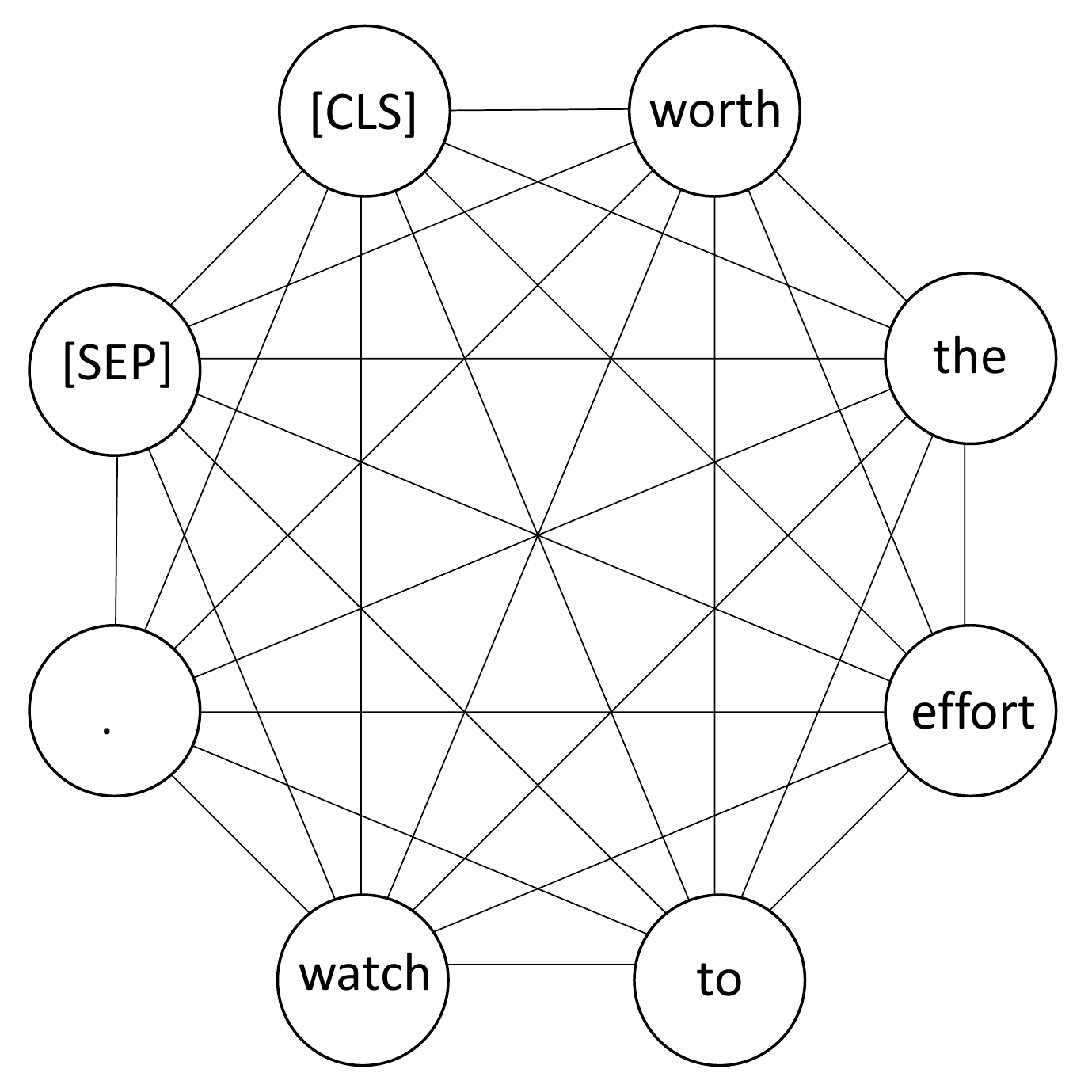}}
\subfigure[Adjacency matrix $\boldsymbol{A}$.\label{subfig:A}]{\includegraphics[width=0.35\textwidth]{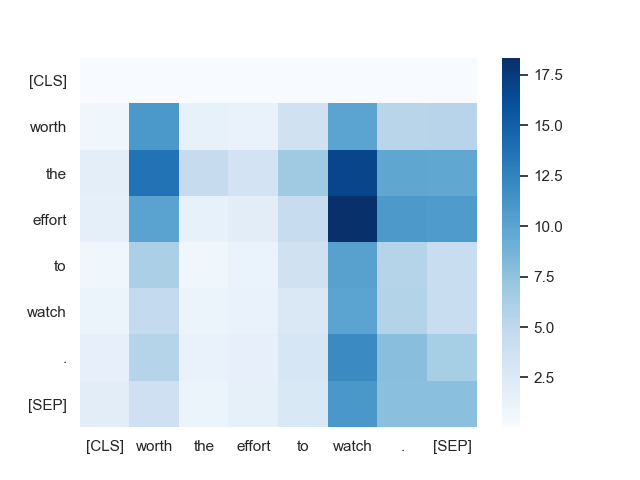}}
\subfigure[Normalized adjacency matrix $\hat{\boldsymbol{A}}$.\label{subfig:hat_A}]{\includegraphics[width=0.35\textwidth]{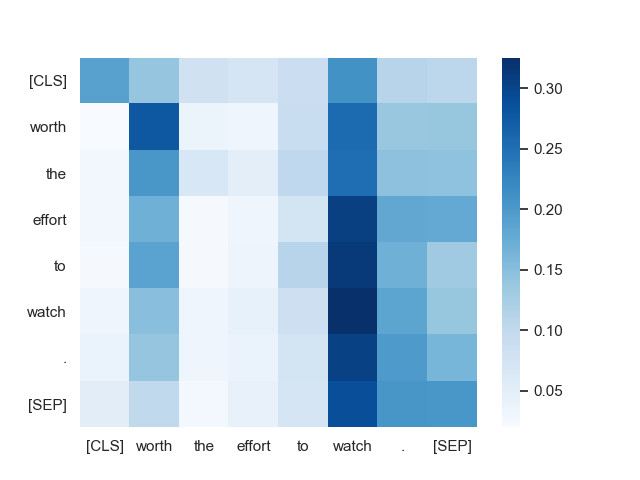}}
\caption{Illustration of self-attention and the corresponding graph
$\mathcal{G}$.
For simplicity,
we drop the self-loops in $\mathcal{G}$.
\label{fig:relationship}}
\end{figure}

Given a Transformer block,
construct a weighted graph $\mathcal{G}$ with
the input tokens
as nodes
and
$\exp(\boldsymbol{Q}_i^\top \boldsymbol{K}_j)$ as the $(i,j)$th entry of its
adjacency matrix $\boldsymbol{A}$.
By
rewriting
the self-attention matrix
$\hat{\boldsymbol{A}}$ in (\ref{eq:matrix})
as
$\hat{A}_{i,j}=\sigma(\boldsymbol{QK}^{\top})_{i,j}=\exp(\boldsymbol{Q}_i^\top
\boldsymbol{K}_{j})/\sum_l\exp(\boldsymbol{Q}_i^\top \boldsymbol{K}_{l})$,
$\hat{\boldsymbol{A}}$
can thus be viewed as
$\mathcal{G}$'s
normalized adjacency matrix
\citep{von2007tutorial}. In other words,
$\hat{\boldsymbol{A}}=\boldsymbol{D}^{-1}\boldsymbol{A}$, where
$\boldsymbol{D}=\text{diag}(d_1, d_2, \dots, d_n)$ and $d_i=\sum_j A_{i,j}$.
Figure~\ref{fig:relationship} shows an example for
the sentence ``worth the effort to watch." from the SST-2 data set
\citep{socher2013recursive} processed by BERT.

Note that graph convolutional network
combined with residual connections (ResGCN) \citep{kipf2017semi} can be expressed as follows.
\begin{equation}
    ResGCN(\boldsymbol{X}) = \boldsymbol{X}+ ReLU(\boldsymbol{D}^{-1/2}\boldsymbol{AD}^{-1/2}\boldsymbol{XW})=\boldsymbol{X}+ ReLU(\hat{\boldsymbol{A}}\boldsymbol{XW}),
\end{equation}
which has the similar form with the self-attention layer in Eq.
(\ref{eq:attn}).
By comparing self-attention module with ResGCN,
we have the following observations:
(i) Since
$A_{i,j}\neq A_{j,i}$ in general, $\mathcal{G}$ in self-attention is a directed graph;
(ii)
$\hat{\boldsymbol{A}}=\boldsymbol{D}^{-1}\boldsymbol{A}$ in self-attention is the random walk
normalization
\citep{chung1997spectral}, while
GCN usually uses
the symmetric normalization version
$\hat{\boldsymbol{A}}=\boldsymbol{D}^{-1/2}\boldsymbol{AD}^{-1/2}$;  (iii) The
attention matrices
constructed
at different Transformer layers
are different, while in
typical graphs,
the adjacency matrices are usually
static.

\subsection{Unshared Attention Matrix vs Shared Attention Matrix \label{sec:ashare}}

As discussed in Section~\ref{sec:over-smooth},
over-smoothing in graph neural networks is
mainly due to
the repeated aggregation operations using the
same adjacency matrix.
To compare the
self-attention matrices ($\hat{\boldsymbol{A}}$'s) at different Transformer layers,
we first flatten the multi-head attention and then measure the cosine similarity
between $\hat{\boldsymbol{A}}$'s
at successive
layers.
Experiment
is performed
with BERT \citep{devlin2019bert}, RoBERTa
\citep{liu2019roberta} and ALBERT \citep{lan2019albert} on the WikiBio data
set
\citep{lebret2016neural}.

\begin{wrapfigure}{r}{7cm}
\includegraphics[width=0.48\columnwidth]{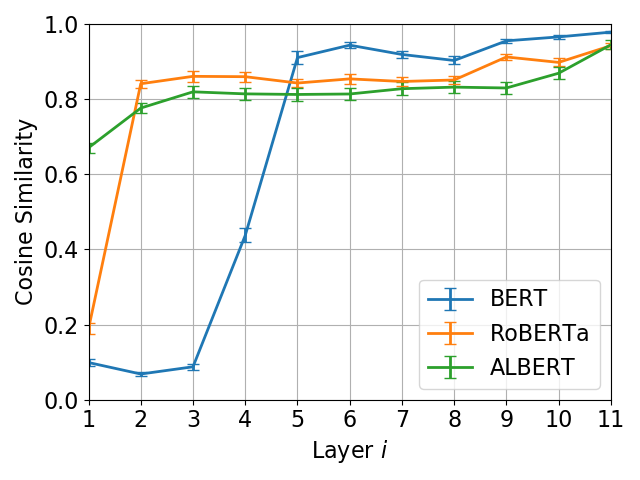}
\caption{Consine similarity between the attention matrices
$\hat{\boldsymbol{A}}$'s
at layer $i$ and its next higher
layer.\label{fig:sim1}}
\end{wrapfigure}

Figure~\ref{fig:sim1}
shows
the cosine similarities obtained. As can be seen,
the similarities
at the last few
layers are high,\footnote{For example,
in BERT,
the attention matrices
$\hat{\boldsymbol{A}}$'s for the last $8$ layers are very similar.
}
while those at the first few layers are different from each other.
In other words,
the attention patterns at the first few layers are changing, and become stable at
the upper layers.

In the following, we
focus on BERT  and
explore how many layers can
share the same
self-attention matrix.
Note that this is different from
ALBERT,
which shares model parameters
instead of attention matrices.
Results are shown in
Table~\ref{tab:ashare}.
As can be seen, sharing attention matrices among the last $8$ layers (i.e.,
layers 5-12) does not harm
model performance. This is consistent with the observation in
Figure~\ref{fig:sim1}.
Note that sharing attention matrices not only reduces
the number of parameters
in the self-attention module, but also
makes the model more efficient by
reducing the
computations
during training and inference.
As shown in Table~\ref{tab:ashare}, BERT (5-12) reduces
$44.4\%$ FLOPs in
the self-attention modules compared with the vanilla BERT, while still achieving
comparable
average GLUE scores.

\begin{table}[t]
\caption{Performance
(\%)
on the GLUE development set
by the original BERT (top row) and various BERT variants with different degrees of
self-attention matrix
sharing.
Numbers in parentheses are the
layers
that share
the self-attention matrix
(e.g., BERT (1-12) means that
the $\hat{\boldsymbol{A}}$'s from
layers
1-12
are shared).
The last column shows
the FLOPs in the self-attention modules. \label{tab:ashare}}
\centering
\resizebox{0.99\textwidth}{!}{
\begin{tabular}{lcccccccccc}
\hline
 &  MNLI (m/mm) & QQP & QNLI & SST-2 & COLA & STS-B & MRPC & RTE & Average & FLOPs \tabularnewline
\hline
BERT & 85.4/85.8 & 88.2 & 91.5 & 92.9 & 62.1 & 88.8 & 90.4 & 69.0 & 83.8 & $2.7$G\tabularnewline
BERT (11-12) & 84.9/85.0 & 88.1 & 91.0 & 93.0 & 62.3 & 89.7 & 91.1 & 70.8 & 84.0 & $2.4$G \tabularnewline
BERT (9-12) & 85.3/85.1 & 88.1 & 90.1	& 92.9 & 62.6 & 89.3 & 91.2 & 68.5 & 83.7 & $2.1$G\tabularnewline
BERT (7-12) & 84.2/84.8 & 88.0 & 90.6 & 92.1 & 62.7 & 89.2 & 90.5 & 68.2 & 83.4 & $1.8$G \tabularnewline
BERT (5-12) & 84.0/84.3 & 88.0 & 89.7 & 92.8 & 64.1 & 89.0 & 90.3 & 68.2 & 83.4  & $1.5$G \tabularnewline
BERT (3-12) & 82.5/82.4 & 87.5 & 88.6 & 91.6 & 57.0 & 87.9 & 88.4 & 65.7 & 81.3  & $1.2$G\tabularnewline
BERT (1-12) & 81.3/81.7 & 87.3 & 88.5 & 92.0 & 57.7 & 87.4 & 87.5 & 65.0 & 80.9 & $1.1$G \tabularnewline
\hline
\end{tabular}}
\end{table}

\section{Over-smoothing in BERT}
In this section, we
analyze the over-smoothing problem in BERT
theoretically,
and then verify the result empirically.

\subsection{Theoretical Analysis}

Our analysis is based on matrix projection.
We define a subspace $\mathcal{M}$, in which each row vector of the element in
this subspace is identical.


\begin{definition}
Define $\mathcal{M}:=\{\boldsymbol{Y}\in\mathbb{R}^{n\times d}|\boldsymbol{Y}=\boldsymbol{eC}, \boldsymbol{C}\in\mathbb{R}^{1\times d}\}$ as a subspace in $\mathbb{R}^{n\times d}$, where $\boldsymbol{e}=[1, 1, \dots, 1]^\top\in\mathbb{R}^{n\times1}$, $n$ is the number of tokens and $d$ is the dimension of token representation.
\end{definition}

Each $\boldsymbol{Y}$ in subspace $\mathcal{M}$ suffers from the over-smoothing issue since the representation of each token is $\boldsymbol{C}$, which is the same with each other.
We define the distance between matrix $\boldsymbol{H}\in\mathbb{R}^{n\times d}$
and $\mathcal{M}$ as $d_\mathcal{M}(H):=\min_{\boldsymbol{Y}\in\mathcal{M}} \Vert
\boldsymbol{H}-\boldsymbol{Y}\Vert_F$, where $\Vert\cdot\Vert_F$ is the Frobenius norm. Next, we investigate the distance between the output of layer $l$ and subspace $\mathcal{M}$.
We have the following Lemma.

\begin{lemma} \label{lemma:1}
For self-attention matrix $\hat{\boldsymbol{A}}$, any $\boldsymbol{H},\boldsymbol{B}\in \mathbb{R}^{n\times d}$ and $\alpha_1, \alpha_2 \geq 0$, we have:
\begin{align}
    d_\mathcal{M}(\boldsymbol{HW}) &\leq sd_\mathcal{M}(\boldsymbol{H}), \\
    d_\mathcal{M}(\text{ReLU}(\boldsymbol{H})) &\leq d_\mathcal{M}(\boldsymbol{H}), \\
    d_\mathcal{M}(\alpha_1 \boldsymbol{H}+\alpha_2 \boldsymbol{B}) &\leq \alpha_1 d_\mathcal{M}(\boldsymbol{H}) + \alpha_2 d_\mathcal{M}(\boldsymbol{B}), \\
    d_\mathcal{M}(\hat{\boldsymbol{A}}\boldsymbol{H}) &\leq \sqrt{\lambda_{\max}} d_\mathcal{M}(\boldsymbol{H}) \label{eq:novel},
\end{align}
where $\lambda_{\max}$ is the
largest eigenvalue of
$\hat{\boldsymbol{A}}^\top(\boldsymbol{I}-\boldsymbol{ee}^\top)\hat{\boldsymbol{A}}$
and $s$ is the largest singular value of $\boldsymbol{W}$.
\end{lemma}

\begin{figure}
    \vspace{-1em}
    \centering
    \includegraphics[width=\textwidth]{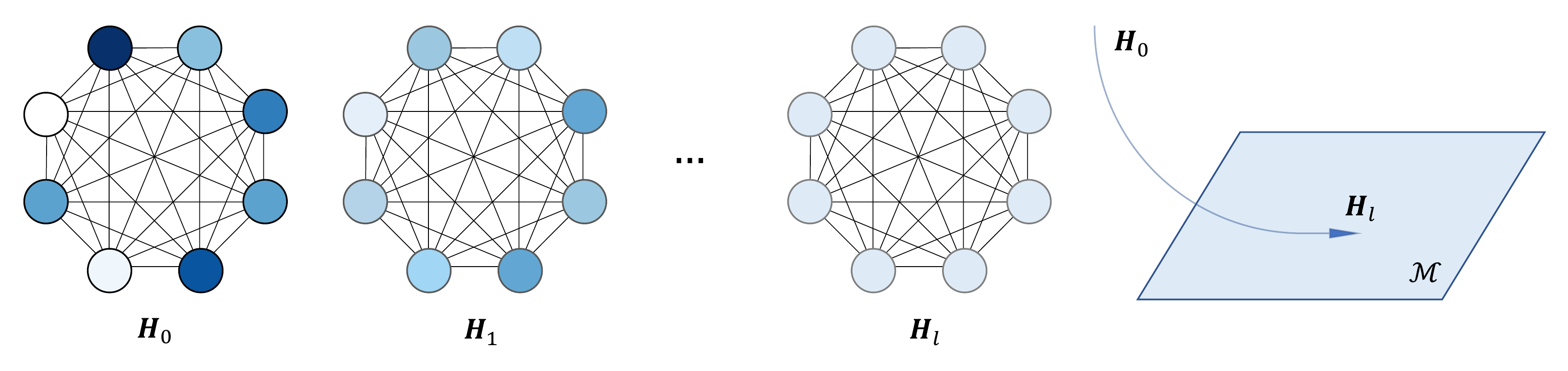}
    \caption{The illustration of over-smoothing problem. Recursively, $\boldsymbol{H}_l$ will converge to subspace $\mathcal{M}$ where representation of each token is identical.}
    \label{fig:illustration}
\end{figure}

Using Lemma~\ref{lemma:1}, we have the following Theorem.
\begin{theorem} \label{theorem:v}
For a BERT block with $h$ heads, we have
\begin{equation}
    d_\mathcal{M}(\boldsymbol{H}_{l+1})\leq vd_\mathcal{M}(\boldsymbol{H}_l),
\end{equation}
where $v=(1+s^2)(1+\sqrt{\lambda} hs)/(\sigma_1\sigma_2)$, $s>0$ is the largest element of all singular values of all $\boldsymbol{W}_l$,
$\lambda$ is the largest eigenvalue of all
$\hat{\boldsymbol{A}}^\top(\boldsymbol{I}-\boldsymbol{ee}^\top)\hat{\boldsymbol{A}}$
for each self-attention matrix $\hat{\boldsymbol{A}}$, and $\sigma_{1}$,
$\sigma_{2}$ are the minimum standard deviation
for two layer normalization operations.
\end{theorem}

Proof is in Appendix~\ref{app:a}.
Theorem~\ref{theorem:v}
shows that
if $v < 1$ (i.e., $\sigma_1\sigma_2>(1+s^2)(1+\sqrt{\lambda} hs)$), the output of layer $l+1$ will be closer to $\mathcal{M}$ than the output of layer $l$.
An illustration of Theorem~\ref{theorem:v}
is shown
in Figure~\ref{fig:illustration}. $\boldsymbol{H}_0$ is the graph
corresponding to the input layer.
Initially,
the
token representations are very different (indicated by the different colors of
the nodes).
Recursively, $\boldsymbol{H}_l$ will converge towards
to subspace $\mathcal{M}$ if $v<1$ and all representations are the same, resulting in over-smoothing.

\textbf{Remark}
Though we only focus on the case $v<1$,
over-smoothing may still exist if $v\geq 1$.

As can be seen, layer normalization plays an important role for the convergence rate $v$.
Interestingly, \cite{dong2021attention} claim that layer normalization plays no
roles for token uniformity,
which seems to conflict with the conclusion in Theorem~\ref{theorem:v}. However,
note that the matrix rank cannot indicate similarity between tokens completely because matrix rank is discrete while similarity is continuous. For instance, given two token embeddings $\boldsymbol{h}_i$ and $\boldsymbol{h}_j$, the
matrix $[\boldsymbol{h}_i, \boldsymbol{h}_j]^\top$ has
rank
$2$ only if $\boldsymbol{h}_i \neq \boldsymbol{h}_j$. In contrast, the consine similarity between tokens is $\frac{\boldsymbol{h}_i^\top \boldsymbol{h}_j}{\Vert\boldsymbol{h}_i\Vert_2\Vert\boldsymbol{h}_j\Vert_2}$.

As discussed in Section~\ref{sec:formulation}, GCN use the symmetric normalization version $\hat{\boldsymbol{A}}=\boldsymbol{D}^{-1/2}\boldsymbol{AD}^{-1/2}$, resulting in the target subspace $\mathcal{M}':=\{\boldsymbol{Y}\in\mathbb{R}^{n\times d}|\boldsymbol{Y}=\boldsymbol{D^{1/2}eC}, \boldsymbol{C}\in\mathbb{R}^{1\times d}\}$ is dependent with adjacent matrix \citep{huang2020tackling}.
In contrast, our subspace $\mathcal{M}$ is
independent of $\hat{\boldsymbol{A}}$ thanks to its random walk normalization.
Thus, Theorem~\ref{theorem:v} can be applied to the vanilla BERT
even though its attention matrix $\hat{\boldsymbol{A}}$ is not similar.

\subsection{Empirical Verification \label{sec:emp}}
Theorem~\ref{theorem:v} illustrates that the magnitude of $\sigma_1\sigma_2$ is important for over-smoothing issue. If $\sigma_1\sigma_2>(1+s^2)(1+\sqrt{\lambda} hs)$, the output will be closer to subspace $\mathcal{M}$ suffered from over-smoothing.
Since $s$ is usually small due to the $\ell_2$-penalty during training
\citep{huang2020tackling}, we neglect the effect of $s$ and compare
$\sigma_1\sigma_2$ with $1$ for simplicity.
To verify the theoretical results, we visualize $\sigma_1\sigma_2$ in different
fine-tuned BERT models. Specifically, we take the development set data of STS-B \citep{cer2017semeval}, CoLA \citep{warstadt2019neural}, SQuAD \citep{rajpurkar2016squad} as input
to the fine-tuned models and visualize the distribution of $\sigma_1\sigma_2$ at the last layer using kernel density estimation \citep{rosenblatt1956remarks}.

\begin{figure}[t]
\vspace{-1.5em}
\centering
\subfigure[STS-B.\label{subfig:var_stsb}]{\includegraphics[width=0.32\textwidth]{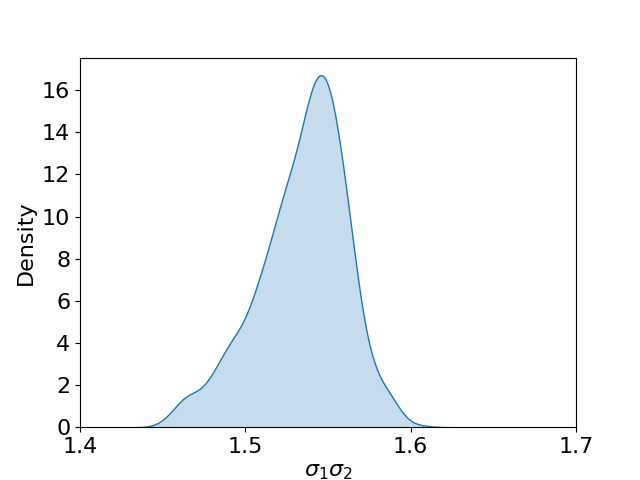}}
\subfigure[CoLA.\label{subfig:var_cola}]{\includegraphics[width=0.32\textwidth]{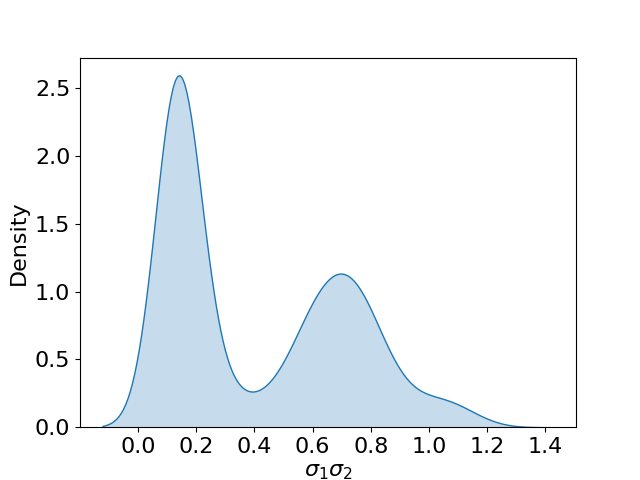}}
\subfigure[SQuAD.\label{subfig:var_squad}]{\includegraphics[width=0.32\textwidth]{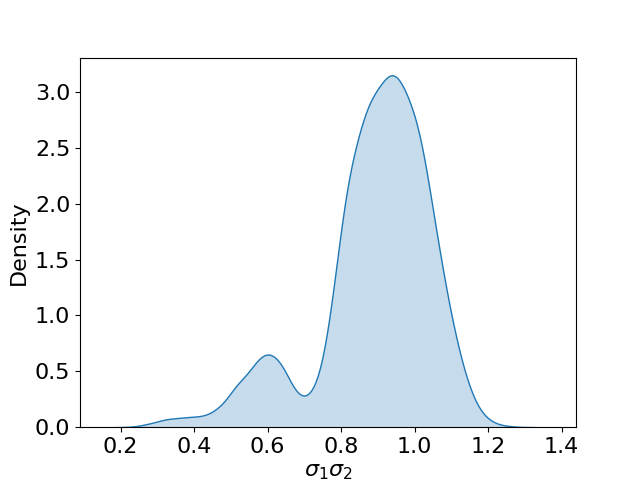}}
\caption{The estimated distribution of $\sigma_1\sigma_2$ for different fine-tuned models.\label{fig:var}}
\vspace{-3mm}
\end{figure}

Results are shown in Figure~\ref{fig:var}.
As can be seen, the distributions of $\sigma_1\sigma_2$ can be very different
across data sets. For STS-B \citep{cer2017semeval}, $\sigma_1\sigma_2$ of all
data is larger than $1$, which means that over-smoothing is serious for this data set. For CoLA \citep{warstadt2019neural} and SQuAD
\citep{rajpurkar2016squad}, there also exists a fraction of samples satisfying
$\sigma_1\sigma_2>1$.

\section{Method} \label{sec:method}
From our proof in Appendix~\ref{app:a}, we figure out that the main reason is
the post-normalization scheme in BERT. In comparison, to train a 1000-layer GCN, \cite{li2021training} instead apply pre-normalization with skip connections to ensure $v>1$.
However, the performance of pre-normalization is not better than post-normalization for layer normalization empirically \citep{he2021realformer}. In this section, we preserve the post-normalization scheme and propose a hierarchical fusion strategy to alleviate the over-smoothing issue.
Specifically,
since only deep layers suffer from the over-smoothing issue, we allow the model select representations from
both shallow layers and deep layers as final output.

\subsection{Hierarchical Fusion Strategy}


\textbf{Concat Fusion}
We first consider a simple and direct layer-wise Concat Fusion approach. Considering a $L$-layer model, we first concatenate the representations $\boldsymbol{H}_k$ from each layer $k$ to generate a matrix $[\boldsymbol{H}_1, \boldsymbol{H}_2, \dots, \boldsymbol{H}_L]$ and then apply a linear mapping to generate the final representation $\sum_{k=1}^L\alpha_k\boldsymbol{H}_k$.
Here $\{\alpha_k\}$ are model parameters independent with inputs.
Since this scheme requires preserving feature maps from all layers, the memory cost will be huge as the model gets deep.

\textbf{Max Fusion}
Inspired by the idea of the widely adopted max-pooling mechanism, we construct the final output by taking the maximum value across all layers for each dimension of the representation.
Max Fusion is an adaptive fusion mechanism since the model can dynamically decide the important layer for each element in the representation.
Max Fusion is the most flexible strategy, since it does not require learning any additional parameters and is more efficient in terms of speed and memory.


\textbf{Gate Fusion} Gate mechanism is commonly used for information propagation in natural language processing field \citep{cho2014learning}.  To exploit the advantages from different semantic levels, we propose a vertical gate fusion module, which predicts the respective importance of token-wise representations from different layers and
aggregate
them adaptively.
Given token representations $\{\boldsymbol{H}^{t}_{k}\}$, where $t$ denotes the token index and $k$ denotes the layer index, the final representation for token $t$ is calculated by
$\sum_{k=1}^L I^{t}_k \cdot \boldsymbol{H}^{t}_k$, where
$I^{t}_1, I^{t}_2, \dots, I^{t}_L = \text{softmax}(g(\boldsymbol{H}^{t}_1), g(\boldsymbol{H}^{t}_2), \dots, g(\boldsymbol{H}^{t}_L))$.
Here $L$ is the number of layers and the gate function $g(\cdot)$ is a fully-connected (FC) layer, which relies on the word representation itself in respective layers to predict its importance scores.
The weights of the gate function $g(\cdot)$ are shared across different layers.

Even though Concat Fusion and Max Fusion have been investigated in the graph field \citep{xu2018representation}, their effectiveness for pre-trained language model have not yet been explored. Besides,  since the \textit{layer-wise} Concat Fusion and \textit{element-wise} Max Fusion lack the ability to generate token representations according to  each token's specificity, we further propose the \textit{token-wise} Gate Fusion for adapting fusion to the language scenario.

\subsection{Experiment Results}
The BERT model is
stacked with $12$ Transformer blocks (Section~\ref{sec:trans_block}) with the
following hyper-parameters: number of tokens $n=128$, number of self-attention heads $h=12$, and hidden layer size $d=768$.
As for the feed-forward layer, we set
the filter size $d_{\ff}$ to 3072 as in \cite{devlin2019bert}.
All experiments are performed on NVIDIA Tesla V100 GPUs.

\begin{table}[t]
\vspace{-1em}
\caption{Performance
(in \%)
of the various
BERT variants
on the GLUE development data set. \label{tab:glue}}
\centering
\resizebox{0.99\textwidth}{!}{
\begin{tabular}{lccccccccc}
\hline
 & MNLI (m/mm) & QQP & QNLI & SST-2 & COLA & STS-B & MRPC & RTE & Average\tabularnewline
\hline
BERT & 85.4/85.8 & 88.2 & 91.5 & 92.9 & 62.1 & 88.8 & 90.4 & 69.0 & 83.8 \tabularnewline
 BERT (concat) & 85.3/85.4 & 87.8 & 91.8 & 93.8 & 65.1 & 89.8 & 91.3 & 71.1& 84.6 \tabularnewline
 BERT (max) & 85.3/85.6 & 88.5 & 92.0 & 93.7 & 64.6 & 90.3 & 91.7 & 71.5 & 84.7 \tabularnewline
 BERT (gate) & 85.4/85.7 & 88.4 & 92.3 & 93.9 & 64.0 & 90.3 & 92.0 & 73.9 & \textbf{85.1} \tabularnewline
\hline
ALBERT & 81.6/82.2 &85.6 & 90.7 & 90.3 & 50.8 & 89.4 & 91.3 & 75.5 & 81.8 \tabularnewline
ALBERT (concat) & 82.8/82.8 & 86.7 & 90.9 & 90.7 & 48.7 & 89.7 & 91.5 & 76.5 & 82.3 \tabularnewline
ALBERT (max)& 82.5/82.8 &86.9& 91.1& 90.7& 50.5 & 89.6 & 92.6 & 77.3& 82.6 \tabularnewline
ALBERT (gate) & 83.0/83.7 & 87.0&  90.9 & 90.4 & 51.3 & 90.0 & 92.4 & 76.2 & \textbf{82.7} \tabularnewline
\hline
\end{tabular}}
\end{table}

\subsubsection{Data and settings}
\textbf{Pre-training}
For the setting in pre-training phase, we mainly follows BERT paper~\citep{devlin2019bert}. Our pre-training tasks are vanilla masked language modeling (MLM) and next sentence prediction (NSP). The pre-training datasets are English BooksCorpus \citep{zhu2015aligning} and Wikipedia  \citep{devlin2019bert} ($16$G in total). The WordPiece embedding \citep{wu2016google} and the dictionary containing  $30,000$ tokens in \citep{devlin2019bert} are still used  in  our paper. To pre-process text, we use the special token {\tt [CLS]} as the first
token of each sequence and {\tt [SEP]}  to separate sentences in a sequence.
The pre-training is performed for $40$ epochs.

\textbf{Fine-tuning} In the fine-tuning phase, we perform downstream experiments on the GLUE \citep{wang2018glue}, SWAG \citep{zellers2018swag} and SQuAD  \citep{rajpurkar2016squad,rajpurkar2018know} benchmarks. GLUE is a natural language understanding benchmark, which includes
three categories tasks:
(i) single-sentence tasks (CoLA and SST-2); (ii) similarity and paraphrase tasks (MRPC, QQP and STS-B); (iii) inference tasks (MNLI, QNLI and RTE).
For MNLI task, we experiment on both the matched (MNLI-m) and mismatched (MNLI-mm) versions. The SWAG data set is for grounded commonsense inference, while SQuAD is a task for question answering.
In SQuAD v1.1 \citep{rajpurkar2016squad}, the answers are included in the context. SQuAD v2.0 \citep{rajpurkar2018know} is more challenge than SQuAD v1.0, in which some answers are not included in the context.
Following BERT \citep{devlin2019bert},  we report accuracy for MNLI, QNLI, RTE, SST-2 tasks, F1 score for QQP and MRPC, Spearman correlation for STS-B, and Matthews correlation for CoLA.
For SWAG task,
we use accuracy for evaluation.
For SQuAD v1.1 and v2.0,
we report the Exact Match (EM) and F1 scores.
Descriptions of the data sets and details of other
hyper-parameter settings are in Appendix~\ref{app:dataset} and in Appendix~\ref{app:hyper}, respectively.

\subsubsection{results and analysis}
\begin{wraptable}{r}{7cm}
\vspace{-1.5em}
\caption{Performance (in \%) on the SWAG and SQuAD development sets.
\label{tab:squad}}
\resizebox{0.49\textwidth}{!}{
\begin{tabular}{lccccc}
\hline
 & SWAG & \multicolumn{2}{c}{SQuAD v1.1} & \multicolumn{2}{c}{SQuAD v2.0} \tabularnewline
 & acc & EM & F1 & EM & F1 \\
\hline
BERT   & 81.6 & 79.7  & 87.1 &  72.9 & 75.5 \tabularnewline
BERT (concat) & 82.0 & 80.2 & 87.8& \textbf{74.1} & 77.0 \tabularnewline
BERT (max)   & 81.9 &80.1 & 87.6 &  73.6 & 76.6 \tabularnewline
BERT (gate) & \textbf{82.1} & \textbf{80.7}  & \textbf{88.0} &  73.9 & \textbf{77.3} \tabularnewline
\hline
\end{tabular}
\label{tab:predictor}}
\vspace{-0.5em}
\end{wraptable}
Since BERT \citep{devlin2019bert} and RoBERTa \citep{liu2019roberta}  share the same architecture and the only difference is data resource and training steps, here we mainly evaluate our proposed method on BERT and ALBERT \citep{lan2019albert}.
Results on the GLUE benchmark are shown in Table~\ref{tab:glue}, while results on SWAG and SQuAD are illustrated in Table~\ref{tab:squad}.
For SQuAD task, in contrast to BERT which \citep{devlin2019bert} utilize the augmented training data during fine-tuning phase, we only fine-tune our model on the standard SQuAD data set. As can be seen, our proposed fusion strategies also perform better than baselines on various tasks consistently.

\begin{figure}[t]
\vspace{-1em}
\centering
\subfigure[STS-B.\label{subfig:stsb}]{\includegraphics[width=0.32\textwidth]{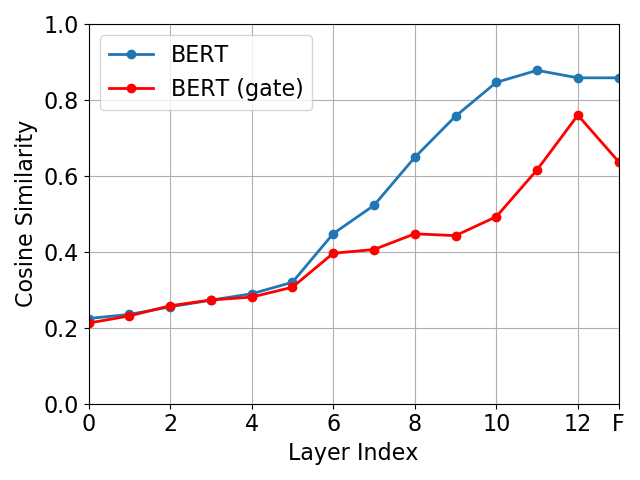}}
\subfigure[CoLA.\label{subfig:cola}]{\includegraphics[width=0.32\textwidth]{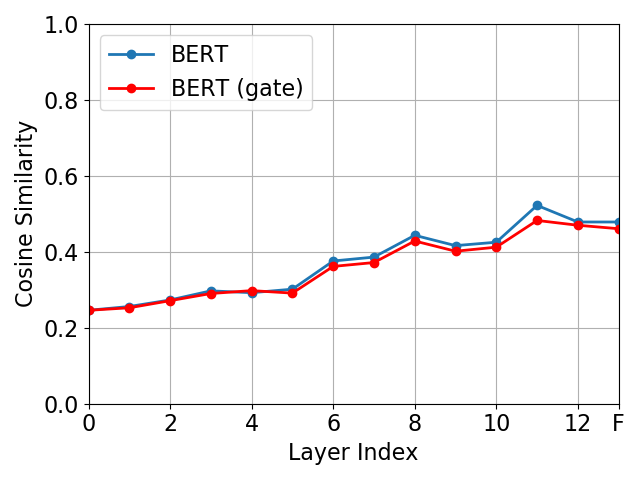}}
\subfigure[SQuAD.\label{subfig:squad2}]{\includegraphics[width=0.32\textwidth]{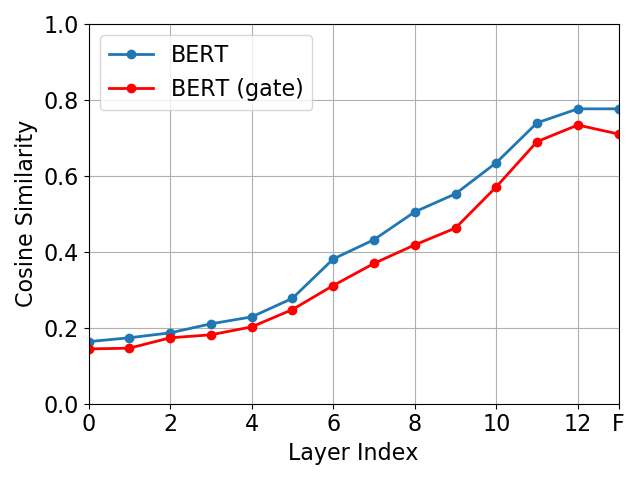}}
\caption{The token-wise similarity comparison between BERT and BERT with gate fusion. Here F means the final output, which is the fusion results for our approach. \label{fig:sim}}
\end{figure}

Following the previous over-smoothing measure, we visualize the token-wise cosine similarity in each layer.
Here we perform visualization on the same data sets as Section~\ref{sec:emp} and the results are shown in Figure~\ref{fig:sim}.
For all three data sets, the cosine similarity has a drop in the last layer compared with baseline. It's remarkable that the similarity drop is the most obvious in STS-B \citep{cer2017semeval}, which is consistent with our empirical verification that STS-B's $\sigma_1\sigma_2$ is the largest in Section~\ref{sec:emp}. Since the representation of tokens from prior layers is not similar with each other, our fusion method alleviates the over-smoothing issue and improve the model performance at the same time.

\begin{figure}[t]
\centering
{\includegraphics[width=0.315\textwidth]{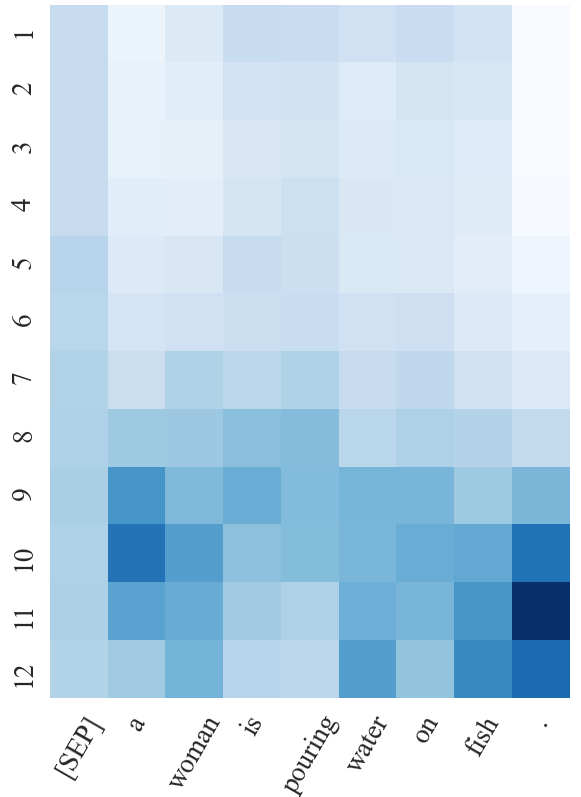}}
{\includegraphics[width=0.26\textwidth]{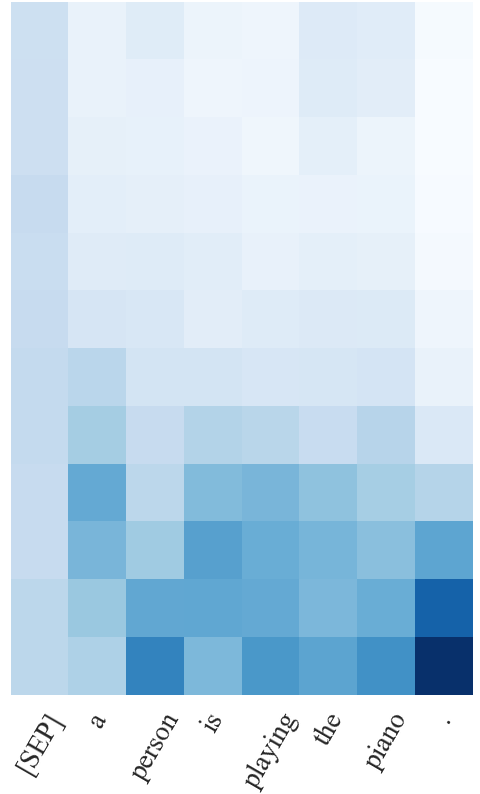}}
{\includegraphics[width=0.363\textwidth]{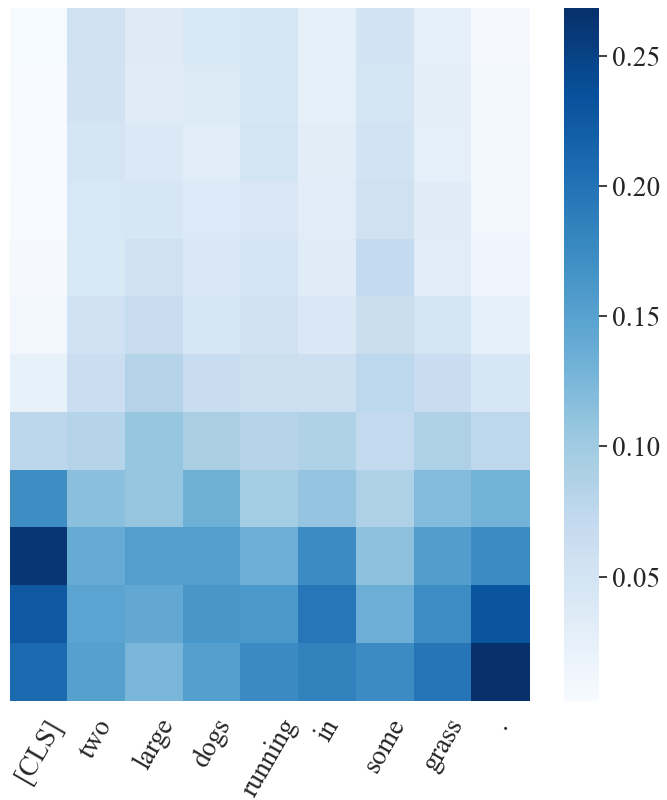}}
\caption{Visualization of importance weights of gate fusion on different layers. \label{fig:weight}}
\end{figure}

To study the dynamic weights of fusion gate strategy, we visualize the importance weight $I_k^t$ for each token $t$ and for each layer $k$. We randomly select three samples and the visualization results are illustrated in Figure~\ref{fig:weight}.
Note that our gate strategy will reduce to vanilla model if representation from the last layer is selected for each token.
As can be seen, the weight distribution of different tokens is adaptively decided, illustrating that the vanilla BERT stacks model is not the best choice for all tokens.  The keywords which highly affect meaning of sentences (i.e. \textit{``women'', ``water'', ``fish''}) are willing to obtain more semantic representations from the deep layer, while for some simple words which appear frequently (i.e. \textit{``a'', ``is''}), the features in shallow layers are preferred.



\section{Conclusion}
In this paper, we revisit the over-smoothing problem in BERT models. Since this issue has been detailed discuss in graph learning field, we firstly establish the relationship between BERT and graph for inspiration, and find out that self-attention matrix can be shared among last few blocks without performance drop. Inspired by over-smoothing discussion in graph convolutional network, we provide some theoretical analysis for BERT models and figure out the importance of layer normalization. Specifically, if the standard derivation of layer normalization is sufficiently large, the output will converge towards to a low-rank subspace. To alleviate the over-smoothing problem, we also propose a hierarchical fusion strategy to combine representations from different layers adaptively. Extensive experiment results on various data sets illustrate the effect of our fusion methods.




\clearpage
\bibliography{iclr2022_conference}
\bibliographystyle{iclr2022_conference}

\clearpage
\appendix
\section{Proof} \label{app:a}
\begin{customlemma}{1}
For self-attention matrix $\hat{\boldsymbol{A}}$, any $\boldsymbol{H},\boldsymbol{B}\in \mathbb{R}^{n\times d}$ and $\alpha_1, \alpha_2 \geq 0$, we have:
\begin{align}
    d_\mathcal{M}(\boldsymbol{HW}) &\leq sd_\mathcal{M}(\boldsymbol{H}), \tag{4} \\
    d_\mathcal{M}(\text{ReLU}(\boldsymbol{H})) &\leq d_\mathcal{M}(\boldsymbol{H}), \tag{5} \\
    d_\mathcal{M}(\alpha_1 \boldsymbol{H}+\alpha_2 \boldsymbol{B}) &\leq \alpha_1 d_\mathcal{M}(\boldsymbol{H}) + \alpha_2 d_\mathcal{M}(\boldsymbol{B}), \tag{6} \\
    d_\mathcal{M}(\hat{\boldsymbol{A}}\boldsymbol{H}) &\leq \sqrt{\lambda_{\max}} d_\mathcal{M}(\boldsymbol{H}) \tag{7} \label{eq:novel},
\end{align}
where $\lambda_{\max}$ is the
largest eigenvalue of
$\hat{\boldsymbol{A}}^\top(\boldsymbol{I}-\boldsymbol{ee}^\top)\hat{\boldsymbol{A}}$
and $s$ is the largest singular value of $\boldsymbol{W}$.
\end{customlemma}
\begin{proof}
Here we only prove the last inequality (\ref{eq:novel}), as the inequity is different from the theories in GCN since $\hat{\boldsymbol{A}}$ is not symmetric and shared in Transformer architecture. For the first three inequalities, we refer to \cite{oono2020graph} and \cite{huang2020tackling}.


Write $\boldsymbol{HH}^\top=\boldsymbol{Q\Omega Q}^\top$ for the eigin-decomposition of $\boldsymbol{HH}^\top$, where $\boldsymbol{Q}=[\boldsymbol{q}_1,\boldsymbol{q}_2, \dots, \boldsymbol{q}_n]$ is the orthogonal and $\boldsymbol{\Omega}=\text{diag}(\omega_1, \dots, \omega_n)$ with all $\omega_i\geq 0$. Recall $\boldsymbol{e}=n^{-1/2}[1,1, \dots, 1]^\top\in\mathbb{R}^{n\times 1}$.

Note that
\begin{align*}
    d_\mathcal{M}(\hat{\boldsymbol{A}}\boldsymbol{H})^2
    &=\Vert(\boldsymbol{I}-\boldsymbol{ee}^\top)\hat{\boldsymbol{A}}\boldsymbol{H} \Vert^2_F \\
    &=tr\{(\boldsymbol{I}-\boldsymbol{ee}^\top)\hat{\boldsymbol{A}}\boldsymbol{HH}^\top\hat{\boldsymbol{A}}^\top(\boldsymbol{I}-\boldsymbol{ee}^\top)\} \\
    &=\sum_{i=1}^n \omega_i\boldsymbol{q}_i^\top\hat{\boldsymbol{A}}^\top(\boldsymbol{I}-\boldsymbol{ee}^\top)\hat{\boldsymbol{A}}\boldsymbol{q}_i.
\end{align*}
Since matrix $\hat{\boldsymbol{A}}^\top(\boldsymbol{I}-\boldsymbol{ee}^\top)\hat{\boldsymbol{A}}$ is positive semidefinite, its all eigenvalues are non-negative. Let $\lambda_{\max}$ be the largest eigenvalue of $\hat{\boldsymbol{A}}^\top(\boldsymbol{I}-\boldsymbol{ee}^\top)\hat{\boldsymbol{A}}$. Consider
\begin{equation*}
    \lambda_{\max}d_\mathcal{M}(\boldsymbol{H})^2-d_\mathcal{M}(\hat{\boldsymbol{A}}\boldsymbol{H})^2=\sum_{i=1}^n\omega_i\boldsymbol{q}_i^\top\{\lambda_{\max}(\boldsymbol{I}-\boldsymbol{ee}^\top)-\hat{\boldsymbol{A}}^\top(\boldsymbol{I}-\boldsymbol{ee}^\top)\hat{\boldsymbol{A}}\}\boldsymbol{q}_i.
\end{equation*}
Let $\boldsymbol{\Sigma}=\lambda_{\max}(\boldsymbol{I}-\boldsymbol{ee}^\top)-\hat{\boldsymbol{A}}^\top(\boldsymbol{I}-\boldsymbol{ee}^\top)\hat{\boldsymbol{A}}$.

Note that $\hat{\boldsymbol{A}}=\boldsymbol{D}^{-1}\boldsymbol{A}$ is a stochastic matrix, we have $\hat{\boldsymbol{A}}\boldsymbol{e}=\boldsymbol{e}$. Thus, $\hat{\boldsymbol{A}}^\top(\boldsymbol{I}-\boldsymbol{ee}^\top)\hat{\boldsymbol{A}}$ has an eigenvalue $0$ and corresponding eigenvecter $\boldsymbol{e}$. Let $\boldsymbol{f}_i$ be a normalised eigenvector of $\hat{\boldsymbol{A}}^\top(\boldsymbol{I}-\boldsymbol{ee}^\top)\hat{\boldsymbol{A}}$ orthogonal to $\boldsymbol{e}$, and $\lambda$ be its corresponding eigenvalue. Then we have
\begin{align*}
    \boldsymbol{e}^\top\boldsymbol{\Sigma e} &= 0, \\
    \boldsymbol{f}_i^\top\boldsymbol{\Sigma f}_i &= \lambda_{\max}-\lambda \geq 0.
\end{align*}
It follows that $d_\mathcal{M}(\hat{\boldsymbol{A}}\boldsymbol{H})^2\leq\lambda_{\max}d_\mathcal{M}(\boldsymbol{H})^2$.
\end{proof}

\textbf{Discussion} Assume further that $\hat{\boldsymbol{A}}$ is doubly stochastic (so that $\hat{\boldsymbol{A}}^\top \boldsymbol{e}=\boldsymbol{e}$) with positive entries. Then by Perron–Frobenius theorem \citep{gantmakher2000theory}, $\hat{\boldsymbol{A}}^\top\hat{\boldsymbol{A}}$ has a maximum eigenvalue $1$ with associated eigenvector $\boldsymbol{e}$ as well.
In this case, the matrix $\hat{\boldsymbol{A}}^\top(\boldsymbol{I}-\boldsymbol{ee}^\top)\hat{\boldsymbol{A}}=\hat{\boldsymbol{A}}^\top\hat{\boldsymbol{A}}-\boldsymbol{ee}^\top$ has a maximum eigenvalue $\lambda_{max}<1$.

\begin{customthm}{2}
For a BERT block with $h$ heads, we have
\begin{equation}
    d_\mathcal{M}(\boldsymbol{H}_{l+1})\leq vd_\mathcal{M}(\boldsymbol{H}_l) \tag{8},
\end{equation}
where $v=(1+s^2)(1+\sqrt{\lambda} hs)/(\sigma_1\sigma_2)$, $s>0$ is the largest element of all singular values of all $\boldsymbol{W}_l$,
$\lambda$ is the largest eigenvalue of all
$\hat{\boldsymbol{A}}^\top(\boldsymbol{I}-\boldsymbol{ee}^\top)\hat{\boldsymbol{A}}$
for each self-attention matrix $\hat{\boldsymbol{A}}$, and $\sigma_{1}$,
$\sigma_{2}$ are the minimum standard deviation
for two layer normalization operations.
\end{customthm}

\begin{proof}
From the definition of self-attention and feed-forward modules, we have
\begin{align*}
    Attn(\boldsymbol{X}) &=\text{LayerNorm}(\boldsymbol{X}+\sum_{k=1}^H \hat{\boldsymbol{A}}^k\boldsymbol{XW}^k+\boldsymbol{1}\boldsymbol{b}^\top) =(\boldsymbol{X}+\sum_{k=1}^H \hat{\boldsymbol{A}}^k\boldsymbol{XW}^k+\boldsymbol{1b}^\top-\boldsymbol{1b}_{LN}^\top)\boldsymbol{D}_{LN}^{-1} \\
    FF(\boldsymbol{X}) &=\text{LayerNorm}(\boldsymbol{X}+\text{ReLU}(\boldsymbol{XW}_1+\boldsymbol{1b_1}^\top)\boldsymbol{W}_2+\boldsymbol{1b}_2^\top)\\
    &=(\boldsymbol{X}+\text{ReLU}(\boldsymbol{XW}_1+\boldsymbol{1b}_1^\top)\boldsymbol{W}_2+\boldsymbol{1b}_2^\top-\boldsymbol{1b}_{LN}^\top)\boldsymbol{D}_{LN}^{-1}
\end{align*}
Based on the Lemma~\ref{lemma:1}, we have
\begin{align*}
    d_\mathcal{M}(Attn(\boldsymbol{X}))
    &=d_\mathcal{M}((\boldsymbol{X}+\sum_{k=1}^h\hat{\boldsymbol{A}}^k\boldsymbol{XW}^k+\boldsymbol{1b}^\top-\boldsymbol{1b}_{LN}^\top)\boldsymbol{D}_{LN}^{-1}) \\
    &\leq d_\mathcal{M}(\boldsymbol{XD}_{LN}^{-1})+d_\mathcal{M}(\sum_{k=1}^h \hat{\boldsymbol{A}}^k\boldsymbol{XW}^k\boldsymbol{D}_{LN}^{-1})+d_\mathcal{M}(\boldsymbol{1}(\boldsymbol{b}-\boldsymbol{b}_{LN})^\top) \\
    &\leq \sigma_1^{-1}d_\mathcal{M}(\boldsymbol{X})+\sum_{k=1}^hd_\mathcal{M}(\hat{\boldsymbol{A}}^k\boldsymbol{XW}^k\boldsymbol{D}_{LN}^{-1}) \\
    &\leq \sigma_1^{-1}d_\mathcal{M}(\boldsymbol{X})+\sqrt{\lambda} h s\sigma_1^{-1}d_\mathcal{M}(\boldsymbol{X}) \\
    &=(1+\sqrt{\lambda} hs)\sigma_1^{-1}d_\mathcal{M}(\boldsymbol{X}). \\
    d_\mathcal{M}(FF(\boldsymbol{X}))
    &= d_\mathcal{M}((\boldsymbol{X}+\text{ReLU}(\boldsymbol{XW}_1+\boldsymbol{1b}_1^\top)\boldsymbol{W}_2+\boldsymbol{1b}_2^\top-\boldsymbol{1b}_{LN}^\top)\boldsymbol{D}_{LN}^{-1})\\
    &\leq d_\mathcal{M}(\boldsymbol{XD}_{LN}^{-1})+d_\mathcal{M}(\text{ReLU}(\boldsymbol{XW}_1+\boldsymbol{1b}_1^\top)\boldsymbol{W}_2\boldsymbol{D}_{LN}^{-1})+d_\mathcal{M}(\boldsymbol{1}(\boldsymbol{b}_2^\top-\boldsymbol{b}_{LN}^\top)\boldsymbol{D}_{LN}^{-1}) \\
    &\leq d_\mathcal{M}(\boldsymbol{XD}_{LN}^{-1})+d_\mathcal{M}(\boldsymbol{XW}_1\boldsymbol{W}_2\boldsymbol{D}_{LN}^{-1})+d_\mathcal{M}(\boldsymbol{1b}_1^\top \boldsymbol{W}_2\boldsymbol{D}_{LN}^{-1}) \\
    &\leq \sigma_2^{-1}d_\mathcal{M}(\boldsymbol{X})+s^2\sigma_2^{-1}d_\mathcal{M}(\boldsymbol{X})\\
    &= (1+s^2)\sigma_2^{-1}d_\mathcal{M}(\boldsymbol{X}).
\end{align*}
It follows that
\begin{equation*}
    d_\mathcal{M}(\boldsymbol{H}_{l+1}) \leq (1+s^2)(1+ \sqrt{\lambda} hs)\sigma_1^{-1}\sigma_2^{-1} d_\mathcal{M}(\boldsymbol{H}_l).
\end{equation*}
\end{proof}

\section{Data Set \label{app:dataset}}
\subsection{MNLI}
The Multi-Genre Natural Language Inference \citep{williams2018broad} is a crowdsourced ternary classification task. Given a premise sentence and a hypothesis sentence, the target is to predict whether the last sentence is an [entailment], [contradiction], or [neutral] relationships with respect to the first one.

\subsection{QQP}
The Quora Question Pairs \citep{chen2018quora} is a binary classification task. Given two questions on Quora, the target is to determine whether these two asked questions are semantically equivalent or not.

\subsection{QNLI}
The Question Natural Language Inference \citep{wang2018multi} is a binary classification task derived from the Stanford Question Answering Dataset \citep{rajpurkar2016squad}. Given sentence pairs (question, sentence), the target is to predict whether the last sentence contains the correct answer to the question.

\subsection{SST-2}
The Stanford Sentiment Treebank \citep{socher2013recursive} is a binary sentiment classification task for a single sentence. All sentences are extracted from movie reviews with human annotations of their sentiment.

\subsection{CoLA}
The Corpus of Linguistic Acceptability \citep{warstadt2019neural} is a binary classification task consisting of English acceptability judgments extracted from books and journal articles. Given a single sentence, the target is to determine whether the sentence is linguistically acceptable or not.

\subsection{STS-B}
The Semantic Textual Similarity Benchmark \citep{cer2017semeval} is a regression task for predicting the similarity score (from $1$ to $5$) between a given sentence pair, whose sentence pairs are drawn from news headlines and other sources.

\subsection{MRPC}
The Microsoft Research Paraphrase Corpus \citep{dolan2005automatically} is a binary classification task. Given a sentence pair extracted from online news sources, the target is to determine whether the sentences in the pair are semantically equivalent.

\subsection{RTE}
The Recognizing Textual Entailment \citep{bentivogli2009fifth} is a binary entailment classification task similar to MNLI, where [neutral] and [contradiction] relationships are classified into [not entailment].

\subsection{SWAG}
The Situations with Adversarial Generations \citep{zellers2018swag} is a multiple-choice task consisting of $113$K questions
about grounded situations. Given a source sentence, the task
is to select the most possible one among four choices for
sentence continuity.

\subsection{SQuAD v1.1}
The Stanford Question Answering Dataset (SQuAD v1.1) \citep{rajpurkar2016squad} is a large-scale question and answer task consisting of $100$K question and answer pairs from more than $500$ articles. Given a passage and the question from Wikipedia, the goal is to determine the start and the end token of the answer text.

\subsection{SQuAD v2.0}
The SQuAD v2.0 task \citep{rajpurkar2018know} is the extension of above SQuAD v1.1, which contains the $100$K questions in SQuAD v1.1 and $50$K unanswerable questions. The existence of unanswerable question makes this task more realistic and challenging.

\section{Implementation Details \label{app:hyper}}
The hyper-parameters of various downstream tasks are shown in Table~\ref{tbl:hyper}.

\begin{table}[h]
\begin{center}
\caption{Hyper-parameters for different downstream tasks. \label{tbl:hyper}}
\resizebox{0.8\textwidth}{!}{
\begin{tabular}{lcccc}
\hline
          & GLUE & SWAG &SQuAD v1.1 & SQuAD v2.0  \tabularnewline
\hline
Batch size & 32 & 16 & 32 & 48  \tabularnewline
Weight decay & [0.1, 0.01]& [0.1, 0.01] &[0.1, 0.01] & [0.1, 0.01] \tabularnewline
Warmup proportion & 0.1 & 0.1 & 0.1& 0.1 \tabularnewline
Learning rate decay & linear & linear &linear & linear \tabularnewline
Training Epochs & 3 & 3 & 3 &2 \tabularnewline
Learning rate & \multicolumn{4}{c}{[2e-5, 1e-5, 1.5e-5, 3e-5, 4e-5, 5e-5]} \tabularnewline
\hline
\end{tabular}}
\end{center}
\label{tab:implementation_details}
\end{table}
\end{document}